\documentclass{article}
\usepackage{ijcai23}

\usepackage{times}
\usepackage{soul}
\usepackage{url}
\usepackage[hidelinks]{hyperref}
\usepackage[utf8]{inputenc}
\usepackage[small]{caption}
\usepackage{graphicx}
\usepackage{amsmath}
\usepackage{amsthm}
\usepackage{booktabs}
\usepackage{algorithm}
\usepackage{algorithmic}
\usepackage{amsfonts}
\usepackage[switch]{lineno}
\newcommand{\std}[1]{\scriptsize{$\pm$#1}}

\usepackage{multirow}
\usepackage{lscape}

\usepackage{bm}
\usepackage{tablefootnote}
\usepackage{url}

\usepackage{lipsum}

\hypersetup{
  colorlinks=true,
  linkcolor=blue,
  citecolor=blue,
  urlcolor=blue
}

\title{Self-Supervised Learning for Time Series: \\
Contrastive or Generative?}

\author{
Ziyu Liu$^1$
\and
Azadeh Alavi$^1$\and
Minyi Li \And
Xiang Zhang$^2$
\affiliations
$^1$RMIT, 
$^2$University of North Carolina, Charlotte\\
\emails
ziyu.liu2@student.rmit.edu.au,
azadeh.alavi@rmit.edu.au\\
minyi.research@gmail.com,
xiang.zhang@uncc.edu
}

\begin{document}

\maketitle

\begin{abstract}
Self-supervised learning (SSL) has recently emerged as a powerful approach to learning representations from large-scale unlabeled data, showing promising results in time series analysis. The self-supervised representation learning can be categorized into two mainstream: contrastive and generative. In this paper, we will present a comprehensive comparative study between contrastive and generative methods in time series.
We first introduce the basic frameworks for contrastive and generative SSL, respectively, and discuss how to obtain the supervision signal that guides the model optimization. We then implement classical algorithms (SimCLR vs. MAE) for each type and conduct a comparative analysis in fair settings. Our results provide insights into the strengths and weaknesses of each approach and offer practical recommendations for choosing suitable SSL methods.
We also discuss the implications of our findings for the broader field of representation learning and propose future research directions.
All the code and data are released at \url{https://github.com/DL4mHealth/SSL_Comparison}.



\end{abstract}

\section{Introduction}
\label{sec:introduction}
The rapid growth of time series data generated by various domains such as finance, healthcare, and environmental monitoring has spurred a demand for robust and efficient deep learning techniques capable of extracting meaningful insights from these vast and complex datasets~\cite{xu2021deep,he2022masked,wen2020time}. 
Although it's relatively easy to generate massive time series data, the data annotating is much more expensive and highly dependent on domain experts~\cite{jiao2022timeautoad}.
Addressing the challenges posed by label scarcity, self-supervised learning (SSL)\footnote{Note, the abbreviation SSL could also be used for `Semi-Supervised Learning' in literature, which is different from this work.} is a promising paradigm to leverage unlabeled data for model training~\cite{henaff2020data}.


Two primary approaches within SSL, contrastive representation learning (such as SimCLR~\cite{chen2020simple}) and generative representation learning (such as Masked Autoencoder~\cite{he2022masked}), have gained significant attention in recent years. Contrastive SSL learns to distinguish between similar and dissimilar data instances by bringing similar data closer and pushing dissimilar data farther apart in the representation space~\cite{yang2022timeclr}. 
On the contrary, generative self-supervised learning grasps the fundamental distribution of the data, thereby enabling the generation of new instances that closely mirror the original data~\cite{he2022masked}. The term 'generative' refers to its capability to create akin data, but the true strength behind this generative property lies in the model's ability to encapsulate and understand the distribution of the data~\cite{goodfellow2020generative,gogna2016semi}.
However, a comprehensive understanding of their relative strengths and weaknesses in the context of time series analysis remains elusive. 

In this paper, we investigate the performance of contrastive and generative SSL methods for time series tasks, aiming to provide a clear comparison and guidelines for their application.
We begin by introducing the fundamental frameworks for both contrastive and generative SSL, with a particular focus on the generation of supervision signals that guide model optimization. Subsequently, we identify a representative model of each approach, implement the respective models, and carry out a comparative analysis under equitable conditions. 
Specifically, we consider SimCLR\footnote{In this work, we adapt SimCLR to time series data by leveraging time-sensitive augmentations. Without specification, the SimCLR mentioned in this work refers to the adapted one instead of the original image-oriented model.} a notable representation of contrastive models~\cite{chen2020simple}, while we perceive the Masked Autoencoder (MAE) as a quintessential example of generative models~\cite{he2022masked}. Both stand as classic, pioneering, and impactful in their respective fields.

This paper bridges a critical research gap through an in-depth comparison of contrastive and generative SSL methods within the scope of time series analysis. The principal contributions of this work are twofold. Firstly, we provide pragmatic guidance for choosing the most fitting SSL approach tailored to distinct problem requisites, thereby augmenting the efficiency of model deployment. Secondly, by discerning the limitations inherent in both SSL techniques, our study lights the path for future research avenues. Furthermore, the insights gleaned from this study possess potential applicability beyond time series, fostering an enriched comprehension of SSL paradigms across diverse domains. Consequently, this work stands as a valuable asset for researchers and practitioners navigating the expanding field of self-supervised learning.

\section{Related Work}
\label{sec:relted_work}
\paragraph{Self-Supervised Learning for Time Series} 
SSL has emerged as a powerful technique for time series analysis addressing challenges associated with limited labeled data or low-quality data annotations. Numerous studies have explored the potential of SSL in various time series applications, such as forecasting, classification, and anomaly detection~\cite{chen2020simple,nonnenmacher2022utilizing,jiao2022timeautoad}. The SSL contains two main streams: contrastive and  generative models~\cite{liu2021self}. The contrastive SSL maps the input sample to representation and then measures the relative distance among similar and dissimilar samples, where minimizing the relative distance serves as a supervision signal to optimize the model. The generative SSL maps an input time series to a representation which is then used to reconstruct the input sample, in which case, the sample itself plays the role of supervision. 



\paragraph{Contrastive SSL}
Contrastive SSL aims to learn representations by comparing positive and negative pairs, thereby encouraging the model to encode meaningful patterns in the data.
Chen et al. introduced SimCLR, a simple framework for contrastive learning of visual representations, which demonstrated state-of-the-art performance on multiple benchmarks~\cite{chen2020simple}. SimCLR was originally proposed for image processing but soon adapted to time series in many studies. 
Franceschi et al. proposed a contrastive predictive coding (CPC) approach to learning representations for multivariate time series, showcasing its effectiveness in time series classification tasks~\cite{henaff2020data}. Zhang et al. developed a Time-Frequency Consistency (TF-C) model to capture the consistency between time domain and frequency domain of time series data to optimize the model and boost the model performance~\cite{zhangself}.

\paragraph{Generative SSL}
Generative SSL is a blossoming area in deep learning earlier than contrastive models. The majority of generative SSL is based on autoencoder and Generative Adversarial Networks (GANs) architecture~\cite{li2023ti,liang2022self}.
One of the pioneering methods in generative SSL is the Variational Autoencoder (VAE)~\cite{kingmaauto}. VAEs use a probabilistic approach to generate new samples by learning the underlying data distribution. VAEs not only capture the complex distribution of the data but also serve as a powerful generative model to produce new data samples. GAN~\cite{goodfellow2020generative} (or semi-supervised GAN) has also shown immense promise in the realm of generative SSL. GANs comprise of two neural networks—a generator and a discriminator—that compete against each other to generate highly realistic data samples. The generator creates data instances with the aim of fooling the discriminator, while the discriminator strives to distinguish between real and fake samples.

Most recently, the Masked Autoencoder for Distribution Estimation (MADE)~\cite{germain2015made} and its extensions, such as PixelCNN~\cite{van2016pixel}, have attracted significant interest. MADE masks the autoencoder's parameters to respect autoregressive properties, enabling it to model complex distributions. It has demonstrated impressive performance in tasks like image and text generation.

The advancements of these generative models have paved the way for more complex SSL models such as Masked Autoencoder (MAE)~\cite{he2022masked}. The MAE incorporates an autoregressive property into an autoencoder, masking some of the inputs during training, thereby forcing the model to predict the masked values, consequently learning the structure of the input data. This approach has proven effective across various domains, including image, text, and speech processing~\cite{zhang2022survey}.

In summary, the field of generative SSL has seen rapid development, with models such as VAE, GANs, MADE, and MAE driving the frontier of innovation and fostering a better understanding of how to extract useful representations from unlabeled data.


\section{Preliminaries and Problem Formulation}
\label{sec:preliminaries}
Here, we provide the basic notation, followed by the problem formulation for SSL in time series.
Let $\mathcal{D}=\{(\boldsymbol{x}_i)| i= 1, 2, \cdots, N\}$ be a dataset of $N$ samples, where $\boldsymbol{x}_i \in \mathbb{R}^{c \times d}$ represents an individual time series sample. The $c$ and $d$ denote the number of channels and the number of timestamps (i.e., length), respectively.
A multivariate time series sample has $c>1$ while univariate time series has $c=1$. 

\noindent\textbf{Self-Supervised Representation Learning.} \textit{A SSL model $f_\theta$ aims to learn a representation $\boldsymbol{z}_i = f_\theta(\boldsymbol{x}_i)$ for a given time series sample $\boldsymbol{x}_i$, where $f_\theta$ is a parametric function with parameters $\theta$. The generated representation $\boldsymbol{z}_i$ can be used for a wide range of downstream tasks such as forecasting, regression, classification, and clustering.
}

In SSL, the learning trajectory is largely dependent on the inherent structure of the data, circumventing the need for explicit labels or annotations. The central premise involves the formulation of auxiliary tasks - for instance, pretext tasks or data reconstruction - that harness the intrinsic structure of the data and can be resolved without direct supervision. By learning to perform these tasks, the model effectively learns the desired data representation.

\begin{figure*}
    \centering
    \includegraphics[width=\textwidth]{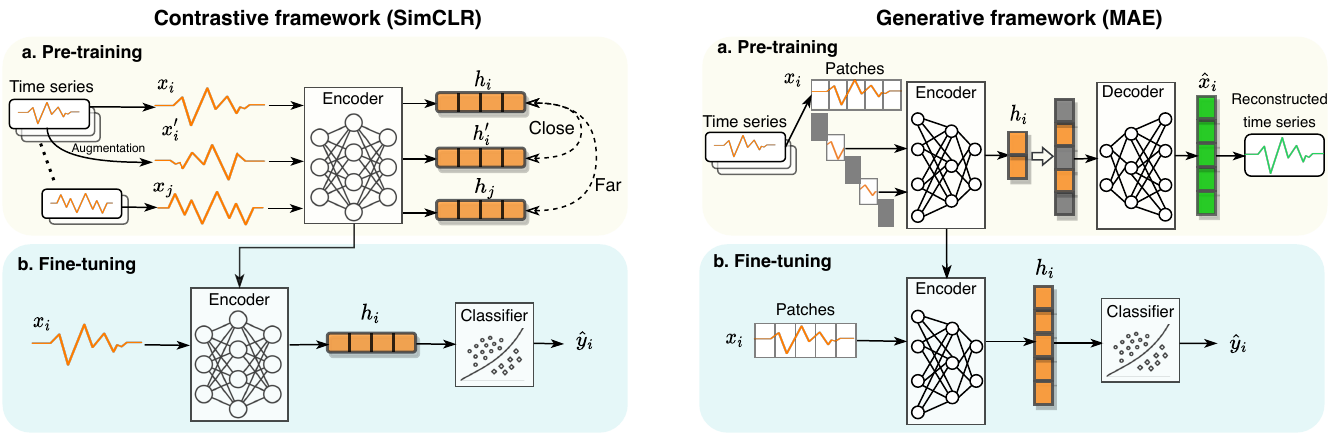}
    \caption{Frameworks of contrastive and generative self-supervised representation learning. The downstream classification task in fine-tuning can be easily extended to other tasks such as forecasting and clustering. In MAE, the gray blocks refer to the masked-out patches.
    }
    \label{fig:frameworks}
\end{figure*}

\section{Contrastive Representation Learning}
\label{sec:contrastive}


\subsection{Overview}
Contrastive representation learning aims to learn robust representations by encouraging the model to distinguish between similar (positive) and dissimilar (negative) pairs of data samples. 
As shown in the left column of Figure~\ref{fig:frameworks}, contrastive SSL first maps the input data to a learned representation space, then employs a contrastive loss function, driving the model to minimize the distance between positive pairings and maximize the distance between negative pairings.
The contrastive learning framework typically consists of two main stages: pre-training and fine-tuning.
Note that the pre-training stage is self-supervised or unsupervised as it does not require labels. But the fine-tuning stage requires sample labels for supervised learning since a downstream classifier is trained at this stage. Therefore, while the term 'self-supervised' in SSL implies label-free representation learning, the overall framework is, in fact, semi-supervised.
It's the same for the generative SSL paradigm.

\subsection{Pre-training}
Pre-training is the most important phase of an SSL model. It generally contains data augmentation, encoding, and loss calculation. For a given time series sample $\boldsymbol{x}_i$, we first create several variants through augmentation. The commonly used time series augmentations include jittering, scaling, permutation, time shifting, slicing and resizing, time masking, frequency masking, neighboring, etc~\cite{liu2023self}. 

In this paper, we simply take jittering, the most popular and effective augmentation, as an example: a noise signal $\boldsymbol{n}_i$ is randomly sampled from a pre-defined distribution (Gaussian distribution is used most often) and added to the original sample $\boldsymbol{x}_i$, resulting in a variant $\boldsymbol{x}'_i$. The basic assumption behind contrastive learning is that $\boldsymbol{x}'_i$ is derived from $\boldsymbol{x}_i$, so they should contain similar information as long as the noise is subtle enough. Likewise, another sample $\boldsymbol{x}_j$ should retain cognate information with its variant $\boldsymbol{x}'_j$.
An encoder will transform all the original samples and augmented samples to a latent space, where we have $\boldsymbol{z}_i = f_\theta(\boldsymbol{x}_i)$, $\boldsymbol{z}'_i = f_\theta(\boldsymbol{x}'_i)$, $\boldsymbol{z}_j = f_\theta(\boldsymbol{x}_j)$, and $\boldsymbol{z}'_j = f_\theta(\boldsymbol{x}'_j)$. Through encoder optimization, we should be able to find a latent space, in which $\boldsymbol{z}'_i$ is closer to $\boldsymbol{z}_i$ compared to $\boldsymbol{z}_j$ and $\boldsymbol{z}'_j$. 

\subsection{Contrastive loss function}
For the above original and augmented samples, we call $(\boldsymbol{x}_i, \boldsymbol{x}'_i)$ as a positive pair while $(\boldsymbol{x}_i, \boldsymbol{x}_j)$ and $(\boldsymbol{x}_i, \boldsymbol{x}'_j)$ are negative pairs.
A contrastive loss is designed to enforce the relative distance among positive and negative pairs. InfoNCE and NT-Xent loss functions are the most classic ones. Take NT-Xent as an example: 
\begin{equation}
    \mathcal{L}_{\textrm{NT-Xent}} = -\frac{1}{N} \sum_{i=1}^{N} \log \frac{\exp\left( \text{sim}(f_\theta(\boldsymbol{x}_i), f_\theta(\boldsymbol{x}'_i)) / \tau \right)}{\sum_{j=1, j \neq i}^{2N} \exp\left( \text{sim}(f_\theta(\boldsymbol{x}_i), f_\theta(\boldsymbol{x}_j)) / \tau \right)},
\end{equation}
where $\text{sim}$ is a  similarity function like cosine similarity and $\tau$ is a temperature parameter to adjust the scale of similarity. The symbol $N$ signifies the number of samples in the training dataset, or more practically, in a training batch. The term $2N$ is used because all $N$ augmented samples are considered as negative samples, effectively doubling the total count of samples for consideration in the loss calculation.

\subsection{Fine-tuning}

In the fine-tuning stage, the pre-trained encoder is adapted to the specific downstream task using a smaller labeled dataset. Given a dataset $\{(\boldsymbol{x}_i, y_i)|i = 1, 2, \cdots, M\}$ with $M$ labelled time series samples, the model parameters $\theta$ are adjusted with a task-specific supervised loss function $\mathcal{L}_{\text{task}}$. Here, the representations generated by the encoder are fed into a task-specific classifier $g_\phi$ with parameters $\phi$.
The objective is to minimize the discrepancy between the predicted labels $\hat{y}_i = g_\phi(f\theta(\boldsymbol{x}_i))$ and the true labels $y_i$.
After fine-tuning, the encoder and classifier can be used to undertake a specific task for an unseen test sample. 


\section{Generative Representation Learning}
\label{sec:generative}

\subsection{Overview}
Broadly speaking,
all the machine learning methods that try to model the data distribution by 
leveraging unlabeled data can be called generative SSL (the right column of Figure~\ref{fig:frameworks}). 
The most commonly used generative SSL approaches are based on autoencoders~\cite{kingmaauto} or GANs~\cite{goodfellow2020generative}. 
In literature, they might not be named as `self-supervised' but `semi-supervised ' or `unsupervised'. 
The key idea is to encode the input data into a latent space and then decode it back into the original space. If the model can achieve a small loss in such a reconstruction task, 
it means the compressed representation in the latent space contains enough information to recover the input data, 
in other words, it's a low-dimensional representation of the input data. Autoencoder assumes that there is some underlying structure or pattern in the data, which can be learned and leveraged to reconstruct the input from the compressed representation~\cite{ju2015deep,vincent2008extracting}.


\subsection{Pre-training}
The model architecture of different generative SSL methods could be different, such as vanilla autoencoder (AE), denoising AE, varitional AE, masked AE, and GAN. Here we introduce a state-of-the-art Masked Autoencoder (MAE) which received over 1800 citations in the short period between it was proposed (Nov. 2021) and the writing of this paper (May. 2023). MAE was originally proposed to increase the scalability of image processing but was adapted to self-supervised time series analysis. For simplicity, we use the same notations of samples and representations in contrastive and generative SSL approaches.

MAE receives a raw input time series sample $\boldsymbol{x}_i$, and slices it into a series of patches (or subsequences). For example, slice a time series with shape $[3, 200]$ into $20$ patches where each patch has shape $[3, 10]$. Then, it randomly selects a small proportion of the patches (like $25\%$ patches) leading to $5 = 20 \times 25\%$ patches, and regards the $5$ patches as input to the MAE for encoding and reconstruction. 
The key assumption behind MAE is that partial information (i.e., a small proportion of patches) is enough to reconstruct the whole sample if the model can successfully seize the data distribution. 
Compared to other generative models, the fundamental distinction in MAE lies in the fact that the encoder operates on \textit{individual patches} of the input, as opposed to transforming the entire sample. For simplicity, we still use $\boldsymbol{x}_i$ to denote a patch in this sample. 

An encoder $f_{\theta}$ transforms the selected patch to a low-dimensional representation $\boldsymbol{h}_i = f_{\theta} (\boldsymbol{x}_i)$. The $\boldsymbol{h}_i$ is used to reconstruct the original patch through a decoder $r_{\psi}$ parameterized by $\psi$. We denote the reconstructed patch as $\hat{\boldsymbol{x}}_i = r_{\psi} (\boldsymbol{h}_i)$.

\subsection{Genrative loss functions}
The most commonly used reconstruction loss function is to measure the distance between the original patches and reconstructed patches through
\begin{equation}
\mathcal{L}_{\text{recon}} = \frac{1}{N} \sum_{i=1}^{N} \Vert \boldsymbol{x}_i - \hat{\boldsymbol{x}_i}  \Vert^2, \quad \text{where} \quad \hat{\boldsymbol{x}_i} = r_{\psi}(f_{\theta}(\boldsymbol{x}_i)).
\end{equation}
The $\Vert \cdot \Vert$ denotes the Euclidean distance, and the objective is to minimize the reconstruction error. 

\subsection{Fine-tuning}

In the fine-tuning stage, the pre-trained encoder $f_{\theta}$ is inherited to a downstream task while the decoder $r_{\psi}$ is not used. 
For a labelled dataset $\{(\boldsymbol{x}_i, y_i)|i = 1, 2, \cdots, M\}$ with $M$, the encoder maps input sample to $\boldsymbol{h}_i = f_{\theta} (\boldsymbol{x}_i)$ and then goes through a downstream classifier $g_{\phi}$ to produce predicted label $\hat{y}_i = g_{\phi} (\boldsymbol{h}_i)$. A classification loss, like cross-entropy, is used to measure the distance between predicted and true labels. 

\subsection{Difference between contrastive and generative SSL models}
We summarize three key differences between contrastive and generative SSL methods. (1) Generative SSL includes a decoder to reconstruct the original sample while contrastive doesn't need this, making contrastive more lightweight. (2) Most generative models use reconstruction in pertaining, in other words, use the original sample itself to guide the encoder optimization; in comparison, a contrastive SSL can not only employ contrastive mapping as shown in Sec.~\ref{sec:contrastive} but also adopt a wide range of pretext tasks such as predictive coding (autoregressive prediction), neighbor detection, trial discrimination, augmentation category detection~\cite{liu2023self}. The flexible architecture provides higher freedom to further modify and improve a contrastive model. 
(3) The loss functions in contrastive SSL measures the relative similarity among embedding (e.g., cosine similarity) in a latent space and/or a pretext classification loss, while generative loss captures Euclidean distance in the original space. 

\begin{table*}[t]
\caption{Performance comparison between SimCLR and MAE. For each method, with pre-training means we used a well-trained encoder from SimCLR or MAE; without pre-training refers to we fine-tune a random initialized encoder (i.e., don't load the pretained model). The `Ratio' column refers to the label ratio, i.e., the proportion of labels used in fine-tuning stage. }
\label{tab:performance}
\resizebox{\textwidth}{!}{%
\begin{tabular}{@{}ccccccccc@{}}
\toprule
\textbf{Model} & \textbf{Ratio} & \textbf{Pretrain} & \textbf{Accuracy} & \textbf{Precision} & \textbf{Recall} & \textbf{F1 Score} & \textbf{AUROC} & \textbf{AUPRC} \\ \midrule
\multirow{10}{*}{\textbf{SimCLR}} & \multirow{2}{*}{0.01} & w & 0.6508\std{0.0283} & 0.6343\std{0.0258} & 0.6353\std{0.0275} & 0.6184\std{0.0279} & 0.8801\std{0.0165} & 0.6963\std{0.0244} \\
 &  & w/o & 0.6366\std{0.0095} & 0.6241\std{0.0178} & 0.6218\std{0.0091} & 0.6008\std{0.0124} & 0.8604\std{0.0095} & 0.6784\std{0.0102} \\\cmidrule{2-9}
 & \multirow{2}{*}{0.1} & w & 0.8522\std{0.0154} & 0.855\std{0.0159} & 0.8481\std{0.017} & 0.8425\std{0.017} & 0.9748\std{0.0035} & 0.911\std{0.0096} \\
 &  & w/o & 0.8156\std{0.0165} & 0.8146\std{0.0164} & 0.8094\std{0.0174} & 0.8024\std{0.0186} & 0.9638\std{0.0037} & 0.8778\std{0.0131} \\\cmidrule{2-9}
 & \multirow{2}{*}{0.3} & w & 0.885\std{0.0086} & 0.8894\std{0.0107} & 0.8835\std{0.0107} & 0.8778\std{0.0118} & 0.9826\std{0.0015} & 0.9349\std{0.0058} \\
 &  & w/o & 0.875\std{0.0116} & 0.8781\std{0.0105} & 0.8745\std{0.0113} & 0.8673\std{0.0127} & 0.9804\std{0.0024} & 0.9272\std{0.0068} \\\cmidrule{2-9}
 & \multirow{2}{*}{0.5} & w & 0.8908\std{0.0032} & 0.8963\std{0.0022} & 0.8893\std{0.0024} & 0.8861\std{0.003} & 0.9848\std{0.0013} & 0.9388\std{0.0017} \\
 &  & w/o & 0.8948\std{0.0049} & 0.8991\std{0.0058} & 0.8958\std{0.005} & 0.8912\std{0.0049} & 0.9836\std{0.0016} & 0.9386\std{0.0027} \\\cmidrule{2-9}
 & \multirow{2}{*}{1.0} & w & 0.8934\std{0.0117} & 0.9011\std{0.0087} & 0.8956\std{0.0126} & 0.8918\std{0.0116} & 0.9845\std{0.0018} & 0.9384\std{0.0022} \\
 &  & w/o & 0.8944\std{0.004} & 0.9005\std{0.0044} & 0.8963\std{0.0038} & 0.8919\std{0.0038} & 0.9845\std{0.0024} & 0.94\std{0.0054} \\ 
 \midrule \midrule
\multirow{10}{*}{\textbf{MAE}} & \multirow{2}{*}{0.01} & w & 0.7895\std{0.0315} & 0.7953\std{0.0311} & 0.7868\std{0.035} & 0.7772\std{0.0335} & 0.933\std{0.0189} & 0.8388\std{0.0416} \\
 &  & w/o & 0.7365\std{0.0148} & 0.743\std{0.017} & 0.7345\std{0.0154} & 0.7219\std{0.0133} & 0.9213\std{0.0099} & 0.8031\std{0.0077} \\ \cmidrule{2-9}
 & \multirow{2}{*}{0.1} & w & 0.8849\std{0.0106} & 0.8894\std{0.0117} & 0.8847\std{0.0125} & 0.8788\std{0.0125} & 0.9744\std{0.0044} & 0.9264\std{0.0101} \\
 &  & w/o & 0.8472\std{0.0228} & 0.8527\std{0.0212} & 0.8467\std{0.0267} & 0.8392\std{0.025} & 0.9643\std{0.0061} & 0.9014\std{0.0156} \\\cmidrule{2-9}
 & \multirow{2}{*}{0.3} & w & 0.8887\std{0.0025} & 0.8942\std{0.0019} & 0.8898\std{0.0002} & 0.8843\std{0.0006} & 0.9723\std{0.006} & 0.9322\std{0.0087} \\
 &  & w/o & 0.882\std{0.0182} & 0.8839\std{0.019} & 0.8834\std{0.0194} & 0.8769\std{0.0194} & 0.9718\std{0.0052} & 0.9294\std{0.0133} \\\cmidrule{2-9}
 & \multirow{2}{*}{0.5} & w & 0.8858\std{0.0085} & 0.89\std{0.0078} & 0.8881\std{0.0081} & 0.8833\std{0.0079} & 0.9708\std{0.0042} & 0.9326\std{0.0058} \\
 &  & w/o & 0.8717\std{0.0095} & 0.875\std{0.0085} & 0.8728\std{0.013} & 0.8667\std{0.0106} & 0.971\std{0.0033} & 0.9257\std{0.011} \\\cmidrule{2-9}
 & \multirow{2}{*}{1.0} & w & 0.8805\std{0.0115} & 0.8868\std{0.0095} & 0.8837\std{0.0106} & 0.8776\std{0.0118} & 0.9669\std{0.0017} & 0.925\std{0.0034} \\
 &  & w/o & 0.8692\std{0.0213} & 0.872\std{0.023} & 0.8714\std{0.0225} & 0.8651\std{0.0223} & 0.9702\std{0.0047} & 0.9236\std{0.0136} \\
 \bottomrule
\end{tabular}%
}
\end{table*}

\section{Experiments}
\label{sec:experiments}

\subsection{Dataset}
\label{sub:dataset}
We conduct comparative experiments with Human Activity Recognition (HAR) data~\cite{anguita2012human}. The HAR dataset is collected from 30 healthy volunteers engaging in six routine activities, namely walking, ascending stairs, descending stairs, sitting, standing, and laying down. The task is to predict these six daily activities. The data were collected at a sampling frequency of 50 Hz using wearable sensors on a smartphone, which captured 3 channels of triaxial linear acceleration. The dataset contains 10299 samples in total. We split it into pre-training set (58\%, 5881), validation set (1471, 14\%), and test set (2947, 28\%). The fine-tuning set is a \textit{subset} of pre-training set with partial labels. 

Taking the pre-training and fine-tuning stages as a whole, the overall model is a semi-supervised model. We use \textit{label ratio} to denote the proportion of samples that have the true labels, i.e., the ratio between fine-tuning and pre-trianing sets. 
We apply upsampling to make the pre-training set balance, resulting in 5,874 samples in total where each activity associates with 979 samples. Each sample contains 3 channels and lasts for 200 timestamps.

\begin{figure*}[t]
    \centering
    \includegraphics[width=\textwidth]{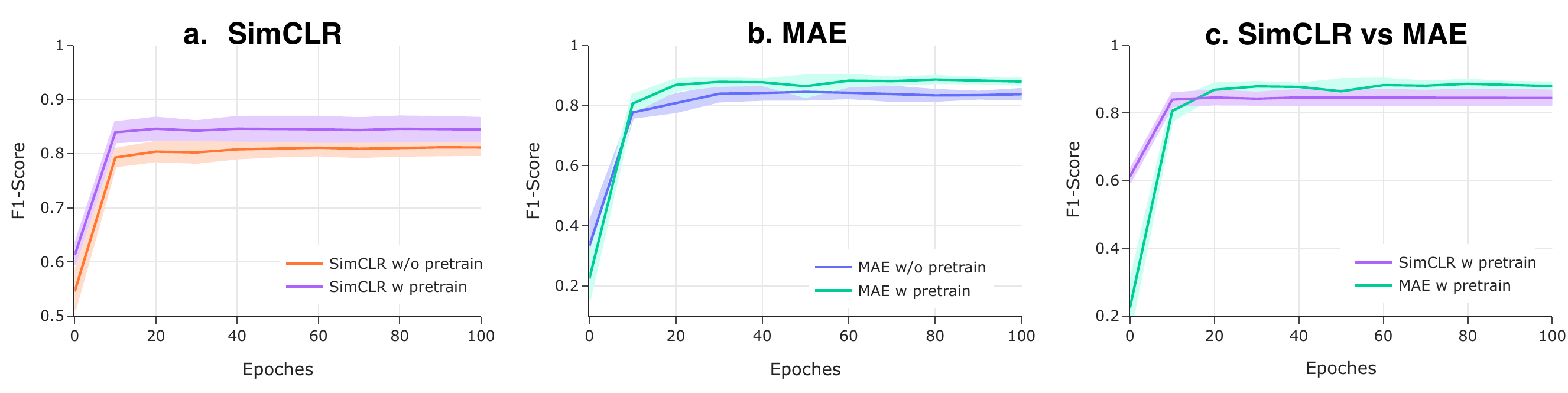}
    \caption{Performance on the test set (label ratio = 0.1; 100 epochs). (a) Comparison of SimCLR's performance with and without pre-training. (b) Comparison of MAE's performance with and without pre-training. (c) A comparative view of SimCLR and MAE, both with pre-training.}
    \label{fig:100epoch}
\end{figure*}

\begin{figure}[t]
    \centering
    \includegraphics[width=0.4\textwidth]{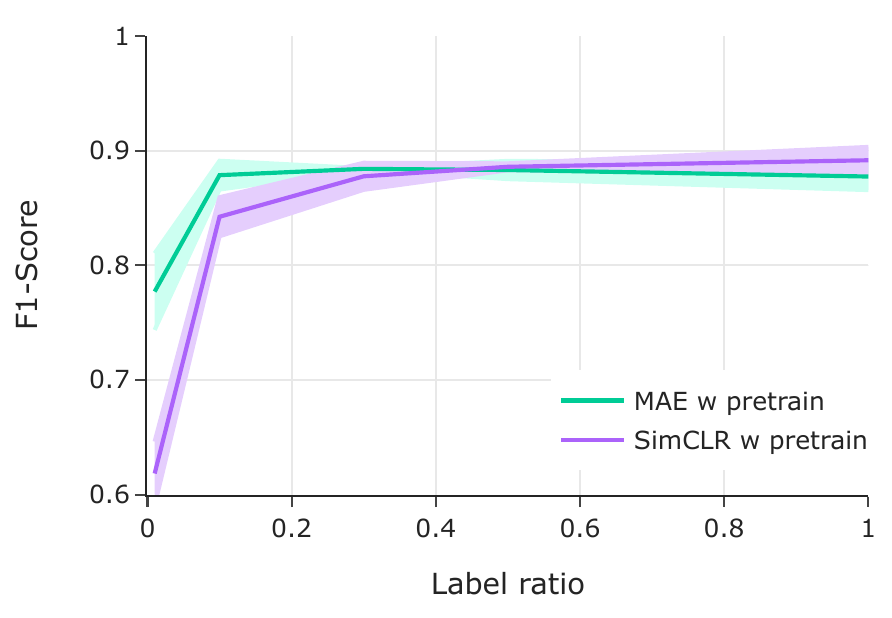}
    \caption{Scalability of SSL Models: Both SSL models demonstrate a significant performance increase when the label ratio escalates from 0.01 to 0.1, and subsequently maintain stability for label ratios exceeding 0.3.}
    \label{fig:label_rato_curve}
\end{figure}

\subsection{Implementation details}
In SimCLR, the encoder employs 2-layer transformer blocks while the classifier has 2 fully-connected layers. We use jittering augmentation and sample the noise from a standard normal Gaussian distribution $\mathcal{N}(0, 1)$. 
The MAE encoder adopts 2 layers of Vision-Transformer (ViT), the decoder is a one-layer ViT, and the fine-tune classifier is also a 2-layer MLP. We select patch size as $[3, 10]$ without overlapping, and mask out 75\% patches in pre-training. The masks' positions are different across samples. All the experimental settings follow the original SimCLR and MAE papers, respectively. For a fair comparison, we set the same hyperparameters in SiCLR and MAE. 
We save the model with the highest F1 score in pre-training (for fine-tuning) and in validation (for testing), respectively. 
The pre-training last for 200 epochs, for each model, and save the model that achieves the lowest loss in pre-training set. 
We run each experiment 5 times using random seeds from 41 to 45, reporting the average and standard deviation. 
We have released all the necessary code and dataset to produce our analysis in \url{https://github.com/DL4mHealth/SSL_Comparison}. 



\subsection{Effectiveness of Pre-training}
We assess whether a pre-training stage contributes to enhancing model performance. Figure~\ref{fig:100epoch} displays the training process over 100 epochs, with a label ratio of 0.1. As can be clearly observed from Figures~\ref{fig:100epoch} (a) and (b), models incorporating pre-training stages surpass those without pre-training by a steady margin of 2\% in terms of F1 score. Even though this margin is not extensive, the consistent performance benefit conferred by pre-training is noteworthy. In summary: 
\begin{itemize}
    \item self-supervised pre-training proves beneficial in building a superior model, regardless of whether the self-supervised learning method is contrastive or generative.
\end{itemize}


\subsection{Overall comparison and scalability} 
\label{sub:finetuning_performance}
To examine the scalability of our method, we observed its performance under various label ratios: 0.01, 0.1, 0.3, 0.5, and 1. We found that the model convergence was slower at a label ratio of 0.01, where all the training dataset was used in pre-training but only 1\% of samples were available in fine-tuning. For such a scenario, we extended the training to 100 epochs (with convergence around the 60th epoch). For other label ratios, we utilized 30 epochs, typically reaching convergence around the 10th to 20th epoch. 

Our comparative study's experimental results are presented in Table~\ref{tab:performance}. From these, we can draw several conclusions:
\begin{itemize}
    \item Across all label ratios, pre-trained models show superior performance compared to those without pre-training.
    \item The lower the label ratio (i.e., fewer labeled samples used in learning), the larger the performance gap. For instance, the pre-trained MAE outperforms its untrained counterpart by 1.25\% in F1 when the label ratio is 1.0. However, this margin increases to 5.53\% when the label ratio drops to 0.01.
    \item With a fixed label ratio of 0.1, as also demonstrated in Figure~\ref{fig:100epoch} (c), SimCLR starts off strong, and converges faster, achieving over 60\% in the initial few epochs. However, MAE eventually reaches a slightly higher level of performance.
    \item Figure~\ref{fig:label_rato_curve} illustrates that MAE outperforms SimCLR when the label ratio is small (e.g., 0.01 and 0.1). However, SimCLR performs slightly better when the label ratio exceeds 0.5. This suggests that MAE is better suited when there's a limited set of labels (fewer than 100 samples per class); for larger label sets, SimCLR should be considered first.
\end{itemize}

\subsection{Training resources}
We have allocated 200 epochs for the pre-training of each SSL model. When tested on the HAR dataset, and utilizing the same hardware, the pre-training time for the MAE was 754 seconds, while SimCLR required 1,016 seconds. This implies that MAE is approximately 25.6\% faster than SimCLR. Hence, if the dataset is substantial and the performance of both models is comparable, we recommend opting for the generative learning approach, in this case, MAE, to save time without compromising on performance.

\subsection{SSL comparison on ECG dataset}

In addition to the HAR dataset, this paper presents ongoing research on the experimental comparison of a classic ECG dataset, specifically the MIT-BIH heart arrhythmia dataset~\cite{moody2001impact}. The dataset comprises 4,000 long-term Holter recordings of 2-lead ECG signals collected from 47 patients. It encompasses 5 distinct classes, and we have balanced the classes through resampling.

Our observations are as follows:

1) The MAE-pre-trained model consistently outperforms the model without pre-training. The performance gap in terms of F1 score increases from 0.8\% to 6.7\% as the label ratio decreases from 50\% to 5\%. Notably, MAE pre-training exhibits an improvement margin of 10.7\% when limited-scale labels (1\%) are available.

2) Conversely, the SimCLR method does not yield a performance boost, as the performance of SimCLR models with and without pre-training is quite similar.

3) However, in the absence of pre-training, SimCLR achieves an F1 score of 83\% with only 1\% labeled data, while MAE only achieves 57\%. We reasonably assume that vanilla SimCLR possesses inherent capabilities to uncover underlying distinctive patterns in small-scale data, thereby leaving limited room for improvement through pre-training. This explains why pre-training is less effective for SimCLR compared to MAE.

4) Interestingly, we observe inconsistent results: SimCLR performs better with 1\% labeled data in the ECG domain, while MAE performs better on the HAR dataset.

To summarize, when working with datasets that contain a small proportion of labeled data, where SSL pre-training is crucial, we recommend employing MAE for human activity data and SimCLR for ECG data. The inconsistent results between different data types may be attributed to variations in data characteristics. The ECG patterns present in this arrhythmia dataset are relatively simple and easily distinguishable, as evidenced by several baseline experiments where even SVM achieves relatively good performance~\cite{liu2021ecg}. Nonetheless, it is important to emphasize that the conclusions drawn above require further validation using additional datasets and SSL models.









\section{Discussion and future work}
\label{sec:discussions}

\subsection{Future work of this study}
While this paper provides a comprehensive comparative study between contrastive and generative SSL methods for time series data, certain limitations within our current study open avenues for future exploration and enhancement.

Firstly, we utilized a single dataset due to space constraints. For a more comprehensive future analysis, it is necessary to involve \textbf{more datasets} through two aspects. Firstly, we should explore a greater variety and diversity of real-world time series datasets. Examples could include ECG, mechanical data, financial data, and climate data. Additionally, the creation of synthetic datasets to analyze how SSL handles time series data with trends, seasonal patterns, and complex structures would be advantageous.

Secondly, the present work confines its comparison to SimCLR and MAE, representing contrastive and generative SSL methods, respectively. Future research could enhance the comparative framework by incorporating \textbf{more state-of-the-art models}. These could include TNC~\cite{tonekaboniunsupervised}, CLOCS~\cite{kiyasseh2021clocs}, TF-C~\cite{zhangself}, and TS2Vec~\cite{yue2022ts2vec} for contrastive learning, along with Semi-VAE~\cite{zhang2019semi}, Semi-GAN~\cite{miao2021generative}, and Denoising AE~\cite{luo2022semi} for generative learning. A broader comparison will offer a more extensive viewpoint on the capabilities and restrictions of various SSL approaches and further enrich our comprehension of their relevance in different contexts.

Furthermore, the focus of our current analysis is predominantly on classification tasks during the fine-tuning stage. Moving forward, we aim to expand the scope of our comparison to encompass \textbf{broader downstream tasks}. These could include forecasting, clustering, and anomaly detection. Such a step will yield a more holistic understanding of contrastive and generative SSL methods' performance across various time series applications.

Lastly, we recommend conducting more detailed analyses of SSL, including visualizing the latent space and examining the transferability of the learned representations. This could potentially uncover further insights into the mechanisms and applications of SSL methodologies.





\subsection{Future work of SSL development}
In contrastive SSL, the selection and size of the negative sample set can notably influence the model efficacy, rendering the identification of an optimal sampling strategy a challenging endeavor. Future work could focus on benchmarking various augmentation techniques and loss functions to ascertain the most effective strategies in different contexts.

In addition, the computational process in generative SSL encompasses a variety of crucial components, such as patch splitting, patch indexing and inverse indexing, in addition to positional encoding. Conversely, contrastive models involve the consideration of a large number of negative samples and the calculation of pairwise distances, which necessitates a training period more than twice as long as that of supervised models. This aspect tends to extend the computational process beyond initial expectations. As a result, future initiatives should emphasize optimizing computational efficiency. This will ensure a more streamlined and efficacious implementation of both generative and contrastive SSL methods.


A promising avenue for future exploration lies in merging generative and contrastive learning approaches within the same model architecture. Such hybrid models could harness the strengths of both methodologies, potentially yielding more robust representation learning for time series data. Future research is therefore encouraged to explore this area, diving into potential synergies between these methods to develop even more effective self-supervised learning models.

\section{Conclusion}
\label{sec:conclusion}
In this paper, we conducted a comprehensive comparative study between contrastive and generative self-supervised learning methods for time series data. 
In general, generative models tend to converge more rapidly and perform impressively when the fine-tuning dataset is quite small (around 100 samples). However, when the dataset is comparatively larger, contrastive models tend to outperform, albeit slightly, their generative counterparts.
Our findings have practical implications for researchers and practitioners working with time series data, as they provide guidance on selecting the most appropriate SSL method for a given scenario. 

\section{Acknowledgments}
This work is partially supported by the National Science Foundation under Grant No. 2245894. Any opinions, findings, conclusions or recommendations expressed in this material are those of the authors
and do not necessarily reflect the views of the funders.

\bibliographystyle{named}
\bibliography{refs}

\end{document}